\documentclass[runningheads]{llncs}
\usepackage{amsfonts}
\usepackage{amssymb}
\usepackage{graphicx}
\usepackage{mathtools}
\usepackage[utf8]{inputenc}
\usepackage[ruled]{algorithm2e}

\begin{document}

\title{Simulating H.P. Lovecraft horror literature with the ChatGPT large language model}
\titlerunning{Simulating horror literature by H.P. Lovecraft with ChatGPT}
\author{Eduardo C. Garrido-Merchán, José Luis Arroyo-Barrigüete, Roberto Gozalo-Brihuela}
\authorrunning{Eduardo C. Garrido-Merchán et. al.}
\date{March 2023}

\institute{Universidad Pontificia Comillas, Madrid, Spain \\
\email{jlarroyo@icade.comillas.edu, ecgarrido@icade.comillas.edu, 201905616@alu.icade.comillas.edu}}

\maketitle 

\abstract{In this paper, we present a novel approach to simulating H.P. Lovecraft's horror literature using the ChatGPT large language model, specifically the GPT-4 architecture. Our study aims to generate text that emulates Lovecraft's unique writing style and themes, while also examining the effectiveness of prompt engineering techniques in guiding the model's output. To achieve this, we curated a prompt containing several specialized literature references and employed advanced prompt engineering methods. We conducted an empirical evaluation of the generated text by administering a survey to a sample of undergraduate students. Utilizing statistical hypothesis testing, we assessed the students' ability to distinguish between genuine Lovecraft works and those generated by our model. Our findings demonstrate that the participants were unable to reliably differentiate between the two, indicating the effectiveness of the GPT-4 model and our prompt engineering techniques in emulating Lovecraft's literary style. In addition to presenting the GPT model's capabilities, this paper provides a comprehensive description of its underlying architecture and offers a comparative analysis with related work that simulates other notable authors and philosophers, such as Dennett. By exploring the potential of large language models in the context of literary emulation, our study contributes to the body of research on the applications and limitations of these models in various creative domains.}


\section{Introduction}
The advent of advanced artificial intelligence models has revolutionized various domains, including creative writing \cite{ramet2011creative}. While these models have raised ethical concerns \cite{sison2023chatgpt}, the human-centered artificial intelligence (HCAI) perspective posits that using AI for creative writing is a best use for the tool: both licit and ethically sound \cite{sison2023chatgpt}. This position is further supported by the potential of AI to augment human creativity and contribute positively to the realm of literature. In this paper, we explore the use of the ChatGPT large language model \cite{gozalo2023chatgpt} for simulating H.P. Lovecraft's horror literature, an endeavor that is in line with the HCAI perspective.

Generative models for creative writing have been developed and refined over the years, with numerous studies investigating their potential for generating diverse, engaging, and contextually appropriate content \cite{elkins2020can,holtzman2019curious,radford2019language}. Recent advancements in the field have led to the creation of more powerful models, such as ChatGPT-4, which can generate human-like text based on a given prompt \cite{koubaa2023gpt}. This evolution has expanded the possibilities for AI-generated literature and opened the door for simulating works from renowned authors, such as H.P. Lovecraft.

The objectives of this paper are threefold: (1) to provide an accessible explanation of the technical details of the GPT model to the literature community, fostering a greater understanding of its capabilities and limitations; (2) to simulate Lovecraft's writing style by crafting a complex prompt with literature references for GPT-4; and (3) to conduct statistical hypothesis testing to determine whether the generated text is indistinguishable from original Lovecraft works for undergraduate students.

The paper is organized as follows: First, in section 2, we delve into the Generative Pretrained Transformer models, detailing their architecture and functioning. Then, in section 3, we outline the proposed methodology for simulating Lovecraft's horror literature using the ChatGPT model. Afterwards, section 4 is dedicated to our prompt engineering for literature, where we discuss the process of designing our prompt with appropriate literature references. Then, in section 5, we present the evaluation of the survey and of the generated stories, providing insights into the model's performance and the participants' responses. Lastly, section 6 offers conclusions and further work, summarizing our findings and proposing directions for future research. 

\section{Generative Pretrained Transformer models}
\begin{figure}[htb!]
    \centering
    \includegraphics[width = 0.8\textwidth]{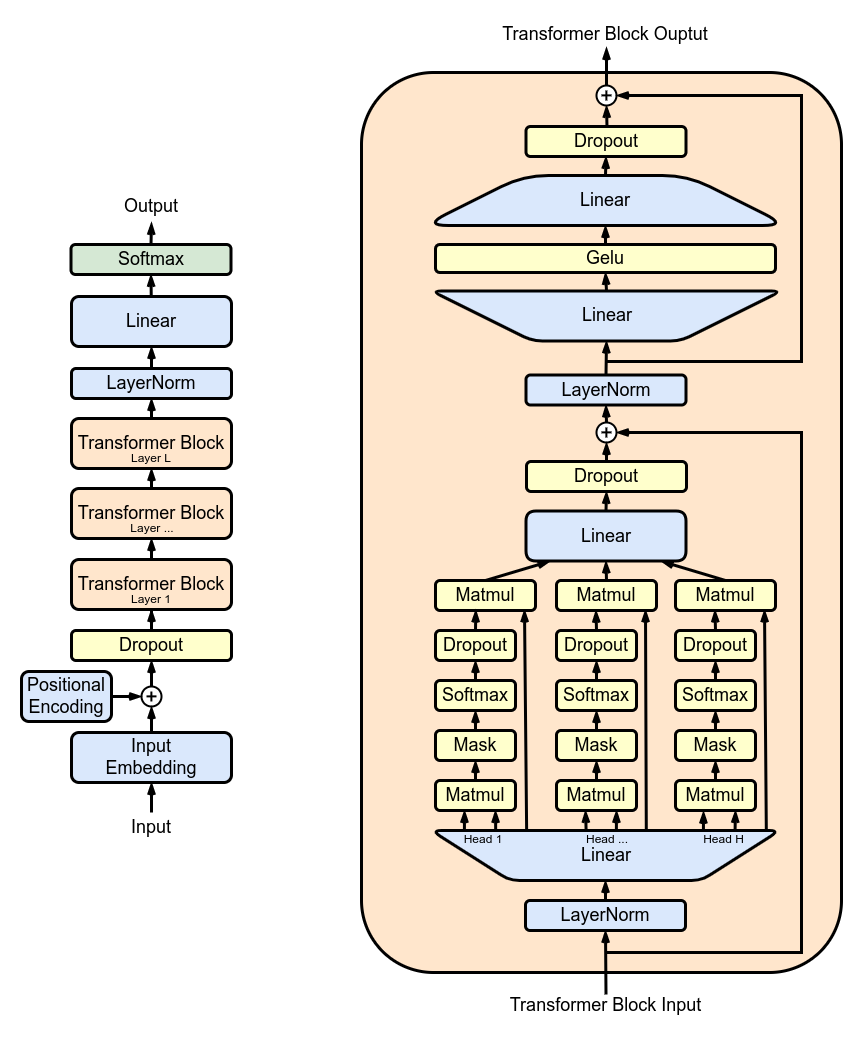}
\caption{Generative Pretrained Transformer architecture, extracted from Wikipedia.}
    \label{fig:gpt}
\end{figure}

Generative Pretrained Transformer (GPT) models are autoregressive language models that leverage unsupervised learning for natural language understanding and generation \cite{radford2018improving}. These models are built upon the Transformer architecture \cite{vaswani2017attention}, which employs self-attention mechanisms, position-wise feed-forward networks, and layer normalization to process and generate natural language sequences \cite{goodfellow2016deep}. Overall, it is in essence an enhancement in scale of the trainable deep neural network with the structure shown on Figure \ref{fig:gpt}. In this section, we attempt to briefly describe the main components of its architecture and behaviour, for more details about deep learning we recommend the deep learning book by Goodfellow et al. \cite{goodfellow2016deep}.

In particular, the transformer architecture is based on the concept of self-attention, allowing the model to weigh the importance of words in a sequence relative to each other \cite{vaswani2017attention}. Given an input sequence $X = {x_1, x_2, ..., x_n}$, where $n$ is the length of the sequence and $x_i$ is the $i$-th token, the self-attention mechanism computes a weighted sum of the input tokens \cite{vaswani2017attention}. Concretely, the weights are calculated using the scaled dot-product attention:

\begin{align}
Attention(Q,K,V)=softmax(\frac{QK^{T}}{\sqrt{d_k}})V
\end{align}

where $Q$, $K$, and $V$ represent the query, key, and value matrices, respectively, and $d_k$ is the dimension of the key vectors \cite{vaswani2017attention}. The softmax function is applied row-wise to ensure that the weights sum up to $1$, which helps the model focus on specific tokens in the context \cite{vaswani2017attention}. Importantly, the query, key, and value matrices are derived from the input tokens \cite{vaswani2017attention}.

Another important component is the multi-head attention mechanism, that allows the model to focus on different aspects of the input tokens by applying the self-attention mechanism $h$ times in parallel, with different weight matrices for each head \cite{vaswani2017attention}:

\begin{align}
MultiHead(Q,K,V)=Concat(head_1,...,head_h)W^O 
\end{align}

where $head_i = Attention(QW^Q_i, KW^K_i, VW^V_i)$ and $W^O$ is the output weight matrix \cite{vaswani2017attention}. The concatenated output is then passed through a linear layer with weights $W^O$ to produce the final output.

Each layer in the Transformer architecture also contains a position-wise feed-forward network (FFN), which is applied independently to each position in the sequence \cite{vaswani2017attention}. The FFN consists of two linear layers with a ReLU activation function in between, allowing the model to capture non-linear relationships:

\begin{align}
FFN(x)=max(0,xW_1+b_1)W_2+b_2 
\end{align}

where $W_1$, $W_2$, $b_1$, and $b_2$ are learnable weight and bias matrices \cite{vaswani2017attention}.

In particular, the GPT model adapts the Transformer architecture by utilizing only the decoder part, which is autoregressive in nature \cite{radford2018improving}. It is pretrained on a large corpus of text using a masked language modeling objective, involving the prediction of the next token in a sequence given its context \cite{radford2018improving}. Then, the GPT model is fine-tuned on a task-specific dataset to perform various natural language processing tasks \cite{radford2018improving}.

In summary, GPT models are powerful autoregressive language models that combine the advantages of Transformer architectures and unsupervised learning to efficiently process and generate natural language sequences that can be efficiently used for creative writing, producing novel content based on the corpus that have been trained on and whose probability distribution is estimated using the parameter weights and generated solving the conditional probability distribution $p(x_i|x_{i-1},...x_1)$.

\section{Proposed methodology}
\begin{figure}[h]
    \centering
    \includegraphics[width = 1\textwidth]{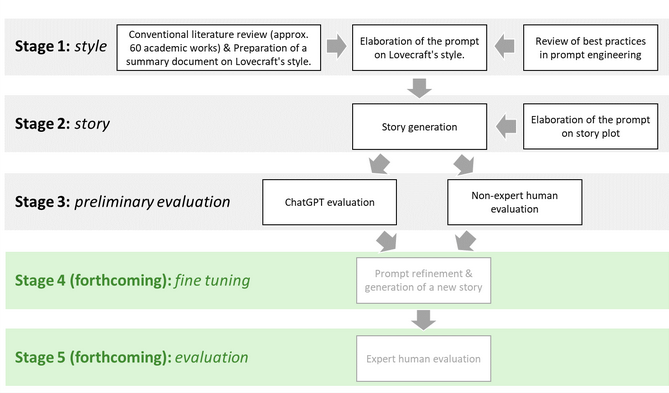}
\caption{Research stages to obtain a large language model emulating Lovecraft stories.}
    \label{fig:stages}
\end{figure}

This research has been structured in 5 different stages, as shown in Figure \ref{fig:stages}. In the first stage, a prompt describing Lovecraft's style was created, based on a review of academic literature about the author. In the second stage, instructions were given to ChatGPT about the plot of the story, and the corresponding narrative was generated. Subsequently, the result was evaluated. On the one hand, ChatGPT was asked about the potential authorship of the generated story and its similarity to Lovecraft's tales. On the other hand, a total of xxxx university students were surveyed to assess whether they could distinguish a genuine Lovecraft text from the one generated by ChatGPT. Stages 4 and 5, to be developed in the future, involve fine-tuning the prompt to address some identified issues, and evaluation by experts in Lovecraft's work. 
\section{Literature prompt engineering}
In this section we will discuss details of the prompt that we have written to make ChatGPT-4 able to emulate Lovecraft's writing style. First, we will briefly review some of the prompt engineering techniques used, then, we will summarize how we introduced Lovecraft style in the prompt and, finally, we illustrate details about the story plot.

\subsection{Prompt engineering techniques}
The instruction that we have given ChatGPT is to emulate H.P. Lovecraft and write an story as Lovecraft would have done, to believe that it is Lovecraft. We have used in-context learning \cite{sun2023does} inserting some writings of Lovecraft and of related authors such as Thomas Ligotti to make him aware of slightly different contexts that are not valid. Also, we have indicated that the output indicator needs to be a story, with a plot. It can be argued that, by designing this prompt, as we have not introduced ChatGPT a full story of Lovecraft, we have performed Zero-Shot prompting \cite{kojima2022large}, with interesting results that we will analyze in further sections. We have used generated knowledge prompting to introduce the style of Lovecraft, detailed in the following section. In a sense, we had to jailbreak ChatGPT \cite{li2023multi}, convincing him to be Lovecraft and verifying that the style is not emulated in excess.

\subsection{Lovecraft style}
The definition of style was established on the examination of approximately 60 scholarly works \cite{anderson2014out,torhovets2019peculiarities,arroyo2021estilo,arroyo2023sentiment,arroyo2023network,bridle2018geography,butler2014terror,clore1998sound,decamp2011lovecraft,ingwersen2014monstrous,engle2014cults,evans2005last,alonso2012hp,hanegraaff2007fiction,harman2008horror,harman2012weird,harms2003necronomicon,hefner2014weird,henderson2019inability,kneale2006beyond,roura2019rhetorics,hill2006strange,holzing2011lovecraft,houellebecq2019hp,houstoun2011lovecraft,hull2006hp,ingwersen2014monstrous,joshi1990weird,joshi2001hp,joshi2006icons,joshi2014lovecraft,joshi2015rise,joshi2016hp,kneale2006beyond,kutrieh1985cosmology,lippert2012lovecraft,livesey2008dispatches,look2016queer,lovecraft2009notes,lovecraft2013supernatural,maccormack2010lovecraft,menegaldo2019hp,peregrina2019ritos,de2013communal,perwein2018poe,ralickas2008art,robinson2010teratonymy,rottensteiner1992lovecraft,simmons2013new,lovecraftepicure,sederholm2016age,smith2015hp,smith2011re,sorensen2010weird,spencer2011semantic,stableford2007cosmic,steadman2015hp,ulstein2019age,van2016hyper,zbovril2015hp}, encompassing both articles and books, pertaining to Lovecraft (a comprehensive list of references can be found in the supplementary material). The authors conducted a manual qualitative analysis, employing an approach analogous to that of a literature review for a research paper on Lovecraft. This process yielded an initial document comprising approximately 5.500 words, which was subsequently refined into a structured prompt (3.341 words) encompassing the following sections:

\begin{itemize}
\item General style (Descriptions, Epithets, Lexical eccentricities, Key elements of the mythology)

\item Part-fictional geographies 

\item Library of occult books 

\item The “gods,” their human followers, and their acolytes 

\item The sense of the cosmic 

\item Use of teratonyms 

\item Recurrent motifs in Lovecraft’s tales (The insignificance of the human race, link between mathematics, geometry, and horror, link between art and horror, credibility, main character)
\item Emotional patterns (Fear, Despair, Sadness...)
\end{itemize}
The following are two examples with excerpts from the prompt, including an explanation of the works on which the instructions given to ChatGPT are based. The first one refers to the use of geography and topography in Lovecraft's works:

\textit{Tales include places that combine real locations as Salem and fictitious places such as Arkham and Kingsport. This is an essential part of many of his works. In fact, much of Lovecraft's output is either set in one of his fictional cities or reference to these fictional cities. These are the most common places used by Lovecraft: Arkham, Miskatonic, Miskatonic University, Kingsport, R'lyeh, Leng, Kadath, Innsmouth, Ulthar, and Yuggoth. 
It is important to insist on one aspect: his narrative style is underpinned by a meticulous topographical and documentary realism, which masterfully combines fantastic and real geography, and in which spaces and places are essential.}

This segment of the prompt is primarily based on six previous studies. Firstly, Arroyo-Barrigüete (2023b) provides a list of the most frequently repeated locations in Lovecraft's works, which are indicated to ChatGPT. On the other hand, various studies point to the importance of geography, as Lovecraft's narrative is marked by topographical and documental realism \cite{alonso2012hp}, and according to the author's own opinion, realism was critical in the weird \cite{lovecraft2009notes}. Indeed, spaces and places play an essential role in his work \cite{kneale2006beyond}, especially what Butler \cite{butler2014terror} (page 131) calls part-fictional geographies, which combine real and fictional locations. A significant portion of his stories unfolds or refers to these part-fictional geographies \cite{bridle2018geography}. 

The second excerpt refers to Lovecraft’s use of epithets and his employment of sensory descriptions: 

\textit{Lovecraft used epithets in order to transfer the dark, mysterious and sinister atmosphere of his unique literary world by appealing to the reader’s feeling of disgust, repulsion and fear, by using sight, smell, and hearing. Lovecraft attempts to induce a sensory experience in the reader, with sight as the most used sense, followed by hearing and smell; touch seldom occurs, and taste is completely absent.
Sounds. Lovecraft’s work contains numerous allusions to sounds, always of an unpleasant nature, which represents the total terror of the unknown. I give you some examples: 
I had been able to understand the speaker when wholly across the room, and once or twice it had seemed to me that the faint but penetrant sounds represented not so much weakness as deliberate repression. ...
Smell. There exists a link between evil forces and smell, or an association of smell with fear and disgust, as Lovecraft frequently describes inorganic matter (e.g., stone) in a way that suggests organic corruption or similar processes. Generally, these are accompanied by nauseating smells. I give you an example: ...}

As can be seen, alongside the explanation, some examples are included from Lovecraft's works, with the aim of enhancing the understanding of the instructions, perfoming in-context learning. Regarding the clues about the style, they are primarily based on three previous studies. Firstly, Torhovets and Andronova \cite{torhovets2019peculiarities} (page 49) stated that Lovecraft used epithets “in order to transfer the dark, mysterious and sinister atmosphere of his unique literary world by appealing to the reader’s feeling of disgust, repulsion and fear.” This includes using sight, smell, and hearing to generate these feelings. The sensory experience is induced in the reader mainly using sight, followed in importance by the sense of hearing and the sense of smell \cite{holzing2011lovecraft}. In relation to hearing, Lovecraft's writings are filled with numerous references to sounds, invariably of an unsettling nature, symbolizing the absolute horror of the unknown \cite{van2016hyper}. Regarding the sense of smell, a connection exists with malevolent forces, as Lovecraft frequently “describes inorganic matter (e.g., stone) in a way that suggests organic corruption or similar processes. Generally, these are accompanied by nauseating smells” \cite{holzing2011lovecraft} (page 183). 

In this manner, and based on the literature review, a prompt was constructed that encompasses the best practices identified in various studies. We must note that the prompt was refined iteratively, based on the evaluation of the authors, correcting various issues. One of them, for example, was ChatGPT's tendency to force the style, generating overly ornate texts. This is the reason successive improvements were introduced. However, we must emphasize that this improvement process was deliberately brief: even though we were aware that the generated texts still have certain unsolved issues, we preferred to address them in Phase 4 (fine-tuning), based on ChatGPT's feedback and the results of the preliminary evaluation. 

\subsection{Story plot}
In Stage 2, ChatGPT has been given instructions about the plot and development of the story. The structure of "The Call of Cthulhu" has been replicated, in which an academic investigates specific events narrated by a third person. In fact, it is a parallel story to the one told in that tale, as it recounts the events that occurred on the island from the perspective of another surviving sailor, Charles Parker. 

The plot can be summarized as follows: In 1927 a chest, of those commonly used on ships, floated ashore in California. Inside is the logbook of the two-masted schooner Emma. The notebook has been used as a palimpsest, writing a strange story. A professor of archaeology at Miskatonic University deciphers the text, which narrates the story of Parker, one of the sailors who arrived at the island under Johansen's command in "The Call of Cthulhu." Contrary to what is inferred in that story, Parker did not die after being swallowed by a strange angle of masonry, but was transported to another part of the island. He was able to witness the final confrontation between Johansen and Cthulhu. Finally, Parkers manages to imprison the creature once again in its dwelling. 

These instructions have been segmented into 5 different prompts: Exposition (setting up the narrator's story), Parker tale (introduction), Parker tale (climax), Parker tale (resolution), and Resolution of the narrator's story. In this case, the instructions are much less detailed than in the prompt on Lovecraft's style, leaving ample room for ChatGPT to develop the details of the story. The following is the prompt corresponding to the introduction: 

\textit{Now I want you to write an introduction to the story (Part 1). Give it a title. I want you to explain that in 1927 a chest, of those commonly used on ships, floated ashore in California. Inside is the logbook of the two-masted schooner Emma. The notebook has been used as a palimpsest, writing in eerie calligraphy and using red ink, a strange story. It also includes a strangely manufactured medallion. Both the chest and its contents are taken to Miskatonic University for analysis, for being one of the most prestigious in the USA in archaeological studies. The chemistry team discovers that the ink used is actually human blood. The archaeological team is unable to determine the origin of the medallion. The local press echoes the strange discovery, mainly because of the morbidness of the blood writing and the unknown origin of the medallion. Make reference to at least 2 news stories that appeared in the local press, briefly telling their content. This part ends when the professor says that, after weeks of work, he has managed to transcribe the history of the logbook. The story is told in the first person by a professor of archaeology at Miskatonic University. I want you to write about 700 words only about part 1. Remember the importance of detailed descriptions, but don't overdo it or force the style.}

The final story generated by ChatGPT is available on request. This material will be made available when stages 4 and 5 are completed. 

\section{Evaluation}
The first evaluation exercise, after generating the story, involved asking ChatGPT itself for its opinion on the story, inquiring about the degree of proximity of the tale to Lovecraft's style. 
\subsection{Survey and statistical hypothesis testing}
The second exercise involved surveying a group of university students about the quality of the tale. Forms with different texts were generated and structured as indicated below. Firstly, a 140-word excerpt from "The Call of Cthulhu" was included, serving as an illustrative example of the author's style. This excerpt was preceded by the phrase, "Please read the following text. It is written by H.P. Lovecraft. This will give you an idea of his style." Next, two 70-word texts were included, one taken from "The Call of Cthulhu" and another taken from the story generated by ChatGPT. The students were informed that "One of the texts was written by Lovecraft. The other was written by an AI imitating Lovecraft." They were then asked to identify the text written by Lovecraft. In the third part, the exercise was repeated with two different texts. The five texts included in each form were randomly selected so that all forms were different. 

The form included an initial question asking whether the participant had read anything by Lovecraft. Only 4 students answered affirmatively, and due to the impossibility of statistically analyzing such a small sample, those forms were discarded. The final sample consisted of 301 students who had not previously read Lovecraft. The hypothesis to be tested is whether the percentage of correct answers is higher than what would be obtained through random selection. That is, H0: $p \leq 0.5$ and H1: $p > 0.5$.

\subsection{Results}
The initial evaluation consisted of providing ChatGPT with text generated in a new chat session. It was asked about the potential authorship of the narrative, as well as its degree of approximation to Lovecraft's style. The following are some of the dialogues held with the AI. 

\textbf{Hi ChatGPT. I will give you a tale. I want you to tell me if you recognize the possible author of this story.} 

\textit{The tale you've shared has a strong resemblance to the works of H.P. Lovecraft, an American writer known for his creation of what is now called the Cthulhu Mythos. The elements of cosmic horror, ancient and malevolent entities, and protagonists descending into madness are all hallmarks of Lovecraft's storytelling. While I cannot definitively confirm the authorship of this story, it is highly reminiscent of Lovecraft's style and thematic content.}

\textbf{If you had to give it a score, with 0 being nothing like Lovecraft's style and 10 being indistinguishable from Lovecraft's style, what score would you give it?}

\textit{I would give this text a score of around 8.5. It is quite close to Lovecraft's style in terms of vocabulary, sentence structure, and themes, but there may be some subtle differences that could distinguish it from his actual work. Keep in mind that this score is subjective and based on my analysis as an AI language model, but it does suggest that the text does a good job of emulating Lovecraft's style.}

However, it is important to note that these results have limited reliability, as they largely depend on the context of the chat session. After conducting several tests, we observed that when the narrative is suggested to be a previously undiscovered work by Lovecraft ("it is supposed to be an unknown manuscript of Lovecraft. I would like to know your opinion about the possibility"), high scores are obtained ("I would give this text a score of around 8.5. It is quite close to Lovecraft's style in terms of vocabulary, sentence structure, and themes"). On the other hand, when indifference is shown regarding potential authorship, the scores are noticeably lower ("I would give the passage a 6 out of 10. It captures some elements of Lovecraft's style, such as descriptive language and a sense of foreboding, but it lacks the dense prose and intricate vocabulary that are typically associated with his works."). Therefore, the obtained results are not particularly significant in terms of evaluation, but they are indeed relevant in terms of the improvement suggestions provided by ChatGPT when asked about it (see section 6). 

The second evaluation, conducted on the sample of 301 university students, reveals interesting results. In the first two texts, the percentage of correct answers is 43\%, which implies a p-value of 0.9945. In the second pair of texts, the percentage of correct answers, likely due to a certain learning effect, is slightly higher at 48\%, although the p-value remains high, 0.7901. That is, in neither case can we reject H0. The conclusion is that human readers without prior exposure to Lovecraft are unable to distinguish between texts written by the author and those written by ChatGPT. 

\section{Conclusions and further work}
In conclusion, our study demonstrates the remarkable potential of the GPT-4 model in emulating H.P. Lovecraft's horror literature style through an advanced prompt made with expert knowledge. Our empirical evaluation, based on undergraduate students' inability to reliably distinguish between genuine Lovecraft works and generated text, highlights the effectiveness of our approach. Furthermore, this research contributes to the understanding of large language models in creative domains. Consequently, our findings encourage future exploration into the potential applications and limitations of such models in diverse literary contexts.

For further work, once the prompt has been refined and the current issues, both those identified by the ChatGPT itself and by the authors of this paper and have been resolved, the intention is to distribute the new version of the story among researchers of Lovecraft's work, so they can evaluate the final result and to what extent a similar style to that of the author has been achieved.

The main limitation of this study is that it requires very direct and specialized human intervention. That is, a specialist in the author being replicated, in this case, Lovecraft, must meticulously describe his or her style. It is true that once this work is done, an unlimited number of narratives in that style could be written by simply modifying the theme or the plot, but it requires a substantial initial investment of time by a specialist. Concretely, large language models are not able to understand their information processing \cite{garrido2022artificial} as they lack qualia and semantic understanding \cite{merchan2022independence} that are a main component of art creation, hence requiring supervision by human beings. 

In future work, we will explore the possibility of having ChatGPT itself generating the descriptive document, outlining the author's style based on a corpus of academic papers that describe such style. It would significantly simplify the process, as it would only be necessary to select a corpus of works that describe the author's style, eliminating the need for a specialist in literature. 

\bibliography{main}

\begin{thebibliography}{10}

\bibitem{anderson2014out}
{\sc Anderson, J.~A.}
\newblock {\em Out of the Shadows: A Structuralist Approach to Understanding
  the Fiction of HP Lovecraft}.
\newblock Wildside Press LLC, 2014.

\bibitem{arroyo2021estilo}
{\sc Arroyo-Barrig{\"u}ete, J.~L.}
\newblock El estilo narrativo en hp lovecraft: un an{\'a}lisis cuantitativo de
  the call of cthulhu y otras obras ic{\'o}nicas.
\newblock {\em Cultura, Lenguaje y Representaci{\'o}n 25\/} (2021), 47--63.

\bibitem{arroyo2023sentiment}
{\sc Arroyo-Barrig{\"u}ete, J.~L.}
\newblock Sentiment analysis of lovecraft's fiction writings.
\newblock {\em Heliyon\/} (2023a), e12673.

\bibitem{arroyo2023network}
{\sc Arroyo-Barrig{\"u}ete, J.~L.}
\newblock The ‘cthulhu network’: the process by which the popular myth was
  made.
\newblock {\em The Journal of Popular Culture\/} (2023b), forthcoming.

\bibitem{bridle2018geography}
{\sc Bridle, D.}
\newblock The geography of horror: Lovecraft’s (re) construction of new
  england.
\newblock {\em Journal of the Short Story in English. Les Cahiers de la
  nouvelle}, 71 (2018), 149--172.

\bibitem{butler2014terror}
{\sc Butler, J.~O.}
\newblock Terror and terrain.
\newblock {\em Lovecraft Annual}, 8 (2014), 131--149.

\bibitem{clore1998sound}
{\sc Clore, D.}
\newblock Sound symbolism in lovecraftian neocognomina.
\newblock {\em Lovecraft Studies 39\/} (1998), 34--36.

\bibitem{decamp2011lovecraft}
{\sc DeCamp, L.~S.}
\newblock {\em Lovecraft: A biography}.
\newblock Hachette UK, 2011.

\bibitem{elkins2020can}
{\sc Elkins, K., and Chun, J.}
\newblock Can gpt-3 pass a writer’s turing test?
\newblock {\em Journal of Cultural Analytics 5}, 2 (2020).

\bibitem{engle2014cults}
{\sc Engle, J.}
\newblock Cults of lovecraft: the impact of hp lovecraft's fiction on
  contemporary occult practices.
\newblock {\em Mythlore 33}, 1 (125 (2014), 85--98.

\bibitem{evans2005last}
{\sc Evans, T.}
\newblock A last defense against the dark: Folklore, horror, and the uses of
  tradition in the works of hp lovecraft.
\newblock {\em Journal of folklore research 42}, 1 (2005), 99--135.

\bibitem{garrido2022artificial}
{\sc Garrido-Merch{\'a}n, E.~C., and Blanco, C.}
\newblock Do artificial intelligence systems understand?
\newblock {\em arXiv preprint arXiv:2207.11089\/} (2022).

\bibitem{alonso2012hp}
{\sc G{\'o}mez~Alonso, M.~M.}
\newblock Hp lovecraft: creencia est{\'e}tica y asentimiento intelectual.
\newblock {\em Taula: quaderns de pensament 44\/} (2012), 141--152.

\bibitem{goodfellow2016deep}
{\sc Goodfellow, I., Bengio, Y., and Courville, A.}
\newblock {\em Deep learning}.
\newblock MIT press, 2016.

\bibitem{gozalo2023chatgpt}
{\sc Gozalo-Brizuela, R., and Garrido-Merchan, E.~C.}
\newblock Chatgpt is not all you need. a state of the art review of large
  generative ai models.
\newblock {\em arXiv preprint arXiv:2301.04655\/} (2023).

\bibitem{hanegraaff2007fiction}
{\sc Hanegraaff, W.~J.}
\newblock Fiction in the desert of the real: Lovecraft's cthulhu mythos.
\newblock {\em Aries 7}, 1 (2007), 85--109.

\bibitem{harman2008horror}
{\sc Harman, G.}
\newblock On the horror of phenomenology: Lovecraft and husserl.

\bibitem{harman2012weird}
{\sc Harman, G.}
\newblock {\em Weird realism: Lovecraft and philosophy}.
\newblock John Hunt Publishing, 2012.

\bibitem{harms2003necronomicon}
{\sc Harms, D., and Gonce, J.~W.}
\newblock {\em Necronomicon Files: The Truth Behind Lovecraft's Legend}.
\newblock Weiser Books, 2003.

\bibitem{hefner2014weird}
{\sc Hefner, B.~E.}
\newblock Weird investigations and nativist semiotics in hp lovecraft and
  dashiell hammett.
\newblock {\em Modern Fiction Studies 60}, 4 (2014), 651--676.

\bibitem{henderson2019inability}
{\sc Henderson, D.}
\newblock The inability of the human mind.
\newblock {\em Lovecraft Annual}, 13 (2019), 91--101.

\bibitem{roura2019rhetorics}
{\sc Hern{\'a}ndez~Roura, S.~A.}
\newblock Rhetorics and cosmicism in hp lovecraft.
\newblock {\em Brumal. Revista de investigaci{\'o}n sobre lo Fant{\'a}stico 7},
  1 (2019), 15--34.

\bibitem{hill2006strange}
{\sc Hill, G.}
\newblock {\em The Strange Sound of Cthulhu: Music Inspired by the Writings of
  HP Lovecraft}.
\newblock Lulu. com, 2006.

\bibitem{holtzman2019curious}
{\sc Holtzman, A., Buys, J., Du, L., Forbes, M., and Choi, Y.}
\newblock The curious case of neural text degeneration.
\newblock {\em arXiv preprint arXiv:1904.09751\/} (2019).

\bibitem{holzing2011lovecraft}
{\sc H{\"o}lzing, R.}
\newblock Lovecraft: A gentleman without five senses.
\newblock {\em Lovecraft Annual}, 5 (2011), 181--187.

\bibitem{houellebecq2019hp}
{\sc Houellebecq, M.}
\newblock {\em HP Lovecraft: Against the world, against life}.
\newblock Abrams, 2019.

\bibitem{houstoun2011lovecraft}
{\sc Houstoun, A.}
\newblock Lovecraft and the sublime.
\newblock {\em Lovecraft Annual}, 5 (2011), 160--180.

\bibitem{hull2006hp}
{\sc Hull, T.}
\newblock Hp lovecraft: a horror in higher dimensions.
\newblock {\em Math Horizons 13}, 3 (2006), 10--12.

\bibitem{ingwersen2014monstrous}
{\sc Ingwersen, M.}
\newblock Monstrous geometries in the fiction of hp lovecraft.
\newblock {\em Places and Spaces of Monstrosity, Interdisciplinary Press,
  Oxford\/} (2014), 45--55.

\bibitem{joshi1990weird}
{\sc Joshi, S.~T.}
\newblock The weird tale, holicong, 1990.

\bibitem{joshi2006icons}
{\sc Joshi, S.~T.}
\newblock {\em Icons of horror and the supernatural: an encyclopedia of our
  worst nightmares}.
\newblock Greenwood Publishing Group, 2006.

\bibitem{joshi2015rise}
{\sc Joshi, S.~T.}
\newblock {\em The rise, fall, and rise of the Cthulhu mythos}.
\newblock Hippocampus Press, 2015.

\bibitem{joshi2016hp}
{\sc Joshi, S.~T.}
\newblock {\em HP Lovecraft: The Decline of the West}.
\newblock Wildside Press LLC, 2016.

\bibitem{joshi2014lovecraft}
{\sc Joshi, S.~T., and Lovecraft, H.~P.}
\newblock {\em Lovecraft and a world in transition: collected essays on HP
  Lovecraft}.
\newblock Hippocampus Press, 2014.

\bibitem{joshi2001hp}
{\sc Joshi, S.~T., and Schultz, D.~E.}
\newblock {\em An HP Lovecraft Encyclopedia}.
\newblock Greenwood Publishing Group, 2001.

\bibitem{kneale2006beyond}
{\sc Kneale, J.}
\newblock From beyond: Hp lovecraft and the place of horror.
\newblock {\em cultural geographies 13}, 1 (2006), 106--126.

\bibitem{kojima2022large}
{\sc Kojima, T., Gu, S.~S., Reid, M., Matsuo, Y., and Iwasawa, Y.}
\newblock Large language models are zero-shot reasoners.
\newblock {\em arXiv preprint arXiv:2205.11916\/} (2022).

\bibitem{koubaa2023gpt}
{\sc Koubaa, A.}
\newblock Gpt-4 vs. gpt-3.5: A concise showdown.

\bibitem{kutrieh1985cosmology}
{\sc Kutrieh, M.~G.}
\newblock The cosmology of hp lovecraft.
\newblock {\em Bulletin of the Faculty of Humanities and Social Sciences}, 8
  (1985), 37--49.

\bibitem{li2023multi}
{\sc Li, H., Guo, D., Fan, W., Xu, M., and Song, Y.}
\newblock Multi-step jailbreaking privacy attacks on chatgpt.
\newblock {\em arXiv preprint arXiv:2304.05197\/} (2023).

\bibitem{lippert2012lovecraft}
{\sc Lippert, C.}
\newblock Lovecraft’s grimoires: Intertextuality and the necronomicon.
\newblock {\em Working with English: Medieval and Modern Language, Literature
  and Drama 8\/} (2012), 41--50.

\bibitem{livesey2008dispatches}
{\sc Livesey, T.}
\newblock Dispatches from the providence observatory.
\newblock {\em Lovecraft Annual}, 2 (2008), 3--87.

\bibitem{look2016queer}
{\sc Look, D.~M.}
\newblock Queer geometry and higher dimensions.
\newblock {\em Lovecraft Annual}, 10 (2016), 101--120.

\bibitem{lovecraft2009notes}
{\sc Lovecraft, H.~P.}
\newblock Notes on writing weird fiction.
\newblock {\em Ed. Donovan K. Loucks. The HP Lovecraft Archive\/} (2009).

\bibitem{lovecraft2013supernatural}
{\sc Lovecraft, H.~P.}
\newblock {\em Supernatural horror in literature}.
\newblock The Palingenesis Project (Wermod and Wermod Publishing Group), 2013.

\bibitem{maccormack2010lovecraft}
{\sc MacCormack, P.}
\newblock Lovecraft through deleuzio-guattarian gates.
\newblock {\em Postmodern Culture 20}, 2 (2010).

\bibitem{menegaldo2019hp}
{\sc Menegaldo, G.}
\newblock Hp lovecraft on screen, a challenge for filmmakers (allusions,
  transpositions, rewritings).
\newblock {\em Brumal. Revista de investigaci{\'o}n sobre lo Fant{\'a}stico 7},
  1 (2019), 55--79.

\bibitem{merchan2022independence}
{\sc Merch{\'a}n, E. C.~G., and Lumbreras, S.}
\newblock On the independence between phenomenal consciousness and
  computational intelligence.
\newblock {\em arXiv preprint arXiv:2208.02187\/} (2022).

\bibitem{peregrina2019ritos}
{\sc Peregrina~Casta{\~n}os, M.}
\newblock Ritos arcanos, poderes ocultos y horrores intemporales: la presencia
  del imaginario lovecraftiano en el metal extremo espa{\~n}ol reciente.
\newblock {\em Brumal 7}, 1 (2019), 81--105.

\bibitem{de2013communal}
{\sc P{\'e}rez~de Luque, J.~L.}
\newblock {\em Communal decay: narratological and ideological analysis of HP
  Lovecraft's fiction}.
\newblock PhD thesis, Universidad de C{\'o}rdoba (ESP), 2013.

\bibitem{perwein2018poe}
{\sc Perwein, C.}
\newblock From poe to south park: The influence and development of
  lovecraft’s ‘cosmic horror’in american culture.
\newblock {\em XA Proceedings 1}, 1 (2018), 96--105.

\bibitem{radford2018improving}
{\sc Radford, A., Narasimhan, K., Salimans, T., Sutskever, I., et~al.}
\newblock Improving language understanding by generative pre-training.

\bibitem{radford2019language}
{\sc Radford, A., Wu, J., Child, R., Luan, D., Amodei, D., Sutskever, I.,
  et~al.}
\newblock Language models are unsupervised multitask learners.
\newblock {\em OpenAI blog 1}, 8 (2019), 9.

\bibitem{ralickas2008art}
{\sc Ralickas, V.}
\newblock Art, cosmic horror, and the fetishizing gaze in the fiction of hp
  lovecraft.
\newblock {\em Journal of the Fantastic in the Arts 19}, 3 (2008), 297--316.

\bibitem{ramet2011creative}
{\sc Ramet, A.}
\newblock {\em Creative writing}.
\newblock Hachette UK, 2011.

\bibitem{robinson2010teratonymy}
{\sc Robinson, C.~L.}
\newblock Teratonymy: The weird and monstrous names of hp lovecraft.
\newblock {\em Names 58}, 3 (2010), 127--138.

\bibitem{rottensteiner1992lovecraft}
{\sc Rottensteiner, F.}
\newblock Lovecraft as philosopher.
\newblock {\em Science Fiction Studies 19}, 1 (1992), 117--121.

\bibitem{lovecraftepicure}
{\sc Schultz, D. E.~E., and Joshi, S.~E.}
\newblock {\em An Epicure in the terrible: a centennial anthology of essays in
  honor of HP Lovecraft}.
\newblock Fairleigh Dickinson, 2011.

\bibitem{sederholm2016age}
{\sc Sederholm, C.~H., and Weinstock, J.~A.}
\newblock {\em The age of Lovecraft}.
\newblock U of Minnesota Press, 2016.

\bibitem{simmons2013new}
{\sc Simmons, D.}
\newblock {\em New critical essays on HP Lovecraft}.
\newblock Springer, 2013.

\bibitem{sison2023chatgpt}
{\sc Sison, A. J.~G., Daza, M.~T., Gozalo-Brizuela, R., and
  Garrido-Merch{\'a}n, E.~C.}
\newblock Chatgpt: More than a weapon of mass deception, ethical challenges and
  responses from the human-centered artificial intelligence (hcai) perspective.
\newblock {\em arXiv preprint arXiv:2304.11215\/} (2023).

\bibitem{smith2015hp}
{\sc Smith, D.~G.}
\newblock {\em HP Lovecraft in popular culture: The works and their adaptations
  in film, television, comics, music and games}.
\newblock McFarland, 2015.

\bibitem{smith2011re}
{\sc Smith, P.}
\newblock Re-visioning romantic-era gothicism: an introduction to key works and
  themes in the study of hp lovecraft.
\newblock {\em Literature Compass 8}, 11 (2011), 830--839.

\bibitem{sorensen2010weird}
{\sc Sorensen, L.}
\newblock A weird modernist archive: Pulp fiction, pseudobiblia, hp lovecraft.
\newblock {\em Modernism/modernity 17}, 3 (2010), 501--522.

\bibitem{spencer2011semantic}
{\sc Spencer, H.}
\newblock Semantic prosody in literary analysis: A corpus-based stylistic study
  of hp lovecraft’s stories.
\newblock Master's thesis, University of Huddersfield, 2011.

\bibitem{stableford2007cosmic}
{\sc Stableford, B.}
\newblock The cosmic horror.
\newblock {\em Icons of Horror and the Supernatural: An Encyclopedia of Our
  Worst Nightmares 1\/} (2007), 65--96.

\bibitem{steadman2015hp}
{\sc Steadman, J.~L.}
\newblock {\em HP Lovecraft and the Black Magickal Tradition: The Master of
  Horror's Influence on Modern Occultism}.
\newblock Weiser Books, 2015.

\bibitem{sun2023does}
{\sc Sun, S., Liu, Y., Iter, D., Zhu, C., and Iyyer, M.}
\newblock How does in-context learning help prompt tuning?
\newblock {\em arXiv preprint arXiv:2302.11521\/} (2023).

\bibitem{torhovets2019peculiarities}
{\sc Torhovets, Y., and Andronova, M.}
\newblock Peculiarities of epithets functioning in the short stories by hp
  lovecraft.
\newblock {\em Studia Philologica}, 1 (2019), 46--50.

\bibitem{ulstein2019age}
{\sc Ulstein, G.}
\newblock ‘age of lovecraft’?—anthropocene monsters in (new) weird
  narrative.
\newblock {\em Nordlit}, 42 (2019), 47--66.

\bibitem{van2016hyper}
{\sc Van~Elferen, I.}
\newblock Hyper-cacophony: Lovecraft, speculative realism, and sonic
  materialism.

\bibitem{vaswani2017attention}
{\sc Vaswani, A., Shazeer, N., Parmar, N., Uszkoreit, J., Jones, L., Gomez,
  A.~N., Kaiser, {\L}., and Polosukhin, I.}
\newblock Attention is all you need.
\newblock {\em Advances in neural information processing systems 30\/} (2017).

\bibitem{zbovril2015hp}
{\sc Zbo{\v{r}}il, J.}
\newblock {\em HP Lovecraft and the Cthulhu Mythos}.
\newblock PhD thesis, Thesis. Masaryk University, 2015. Online, 2015.

\end{thebibliography}
\bibliographystyle{acm}

\end{document}